# Improved mmFormer for Liver Fibrosis Staging via Missing-Modality Compensation

Zhejia Zhang[✉], Junjie Wang, and Le Zhang

School of Engineering, College of Engineering and Physical Sciences,
University of Birmingham, Birmingham, UK
`zhejiazhang2026@u.northwestern.edu`

**Abstract.** In real-world clinical settings, magnetic resonance imaging (MRI) frequently suffers from missing modalities due to equipment variability or patient cooperation issues, which can significantly affect model performance. To address this issue, we propose a multimodal MRI classification model based on the mmFormer architecture with an adaptive module for handling arbitrary combinations of missing modalities. Specifically, this model retains the hybrid modality-specific encoders and the modality-correlated encoder from mmFormer to extract consistent lesion features across available modalities. In addition, we integrate a missing-modality compensation module which leverages zero-padding, modality availability masks, and a Delta Function with learnable statistical parameters to dynamically synthesize proxy features for recovering missing information. To further improve prediction performance, we adopt a cross-validation ensemble strategy by training multiple models on different folds and applying soft voting during inference. This method is evaluated on the test set of Comprehensive Analysis & Computing of REal-world medical images (CARE) 2025 challenge, targeting the Liver Fibrosis Staging (LiFS) task based on non-contrast dynamic MRI scans including T1-weighted imaging (T1WI), T2-weighted imaging (T2WI), and diffusion-weighted imaging (DWI). For Cirrhosis Detection and Substantial Fibrosis Detection on in-distribution vendors, our model obtains accuracies of 66.67%, and 74.17%, and corresponding area under the curve (AUC) scores of 71.73% and 68.48%, respectively.

**Keywords:** 3D Medical Image Analysis, Liver Fibrosis Staging, Multimodal Deep Learning

## 1  Introduction

Liver fibrosis results from the sustained overproduction and accumulation of extracellular matrix (ECM) proteins and is common across chronic liver diseases, such as viral hepatitis, alcohol-related liver disease, and metabolic steatohepatitis [1]. This sustained overaccumulation of ECM causes fibrotic scarring and distortion of the normal liver architecture, while the development of regenerative nodules marks the transition to cirrhosis [1]. Timely and accurate diagnostic and therapeutic strategies are crucial for slowing the progression of liver cirrhosis [2].

Although liver biopsy has been regarded as the "gold standard" for diagnosing liver fibrosis, its limitations—including sampling variability, interpretative subjectivity, risk of complications, and high cost—have driven the development and widespread adoption of non-invasive alternatives such as imaging test and elastography to reduce the reliance on biopsy [3,4]. Notably, multimodal MRI scans have become one of the most effective methods for liver fibrosis assessment, as they can capture complementary diagnostic information [5].

With the development of technology, increasingly advanced methods have been developed and integrated with existing non-invasive diagnostic tools for liver fibrosis assessment. Among these, artificial intelligence (AI) and deep learning have demonstrated significant potential in medical image analysis. Compared to traditional diagnostic approaches, AI-based methods provide not only improved efficiency and accuracy, but also enhanced robustness to external variability, enabling more consistent and reliable decision-making across diverse clinical settings [6].

However, the practical application of AI models has proven to be less straightforward than initially expected. In clinical practice, one or more MRI modalities may be missing due to hardware failures, patient non-cooperation, or inconsistent procedures, which result in the loss of specific information and prevents the model from fully capturing target features, thereby significantly compromising its accuracy and generalizability [5]. To address the challenge of missing MRI modalities, researchers have proposed various strategies, including modality imputation, modality synthesis and feature distillation to improve the performance of models [7-9].

In this study, we improve upon the original mmFormer architecture by replacing the segmentation head with a classification head and incorporating a missing-modality compensation module which generates proxy features to recover missing modality information [7]. To increase robustness while improving generalization performance, we adopt a cross-validation strategy during training and apply soft voting ensemble inference at validation, achieve more stable and reliable classification performance in real-world clinical scenarios.

## 2    Method

### 2.1    Dataset

The non-contrast subtasks of the LiFS challenge are based solely on three non-enhanced MRI modalities: T1WI, T2WI, and DWI, selected from the LiQA dataset of CARE 2025 challenge for the development and evaluation of liver fibrosis staging models [10-12]. This dataset includes MRI scans from 610 patients diagnosed with liver fibrosis, acquired using three different scanner vendors: Philips Ingenia 3.0T, Siemens Skyra 3.0T, and Siemens Aera 1.5T. The dataset is partitioned into 360 cases for training, 60 for validation, and 190 for testing. To assess out-of-vendor generalization, the test set includes 70 cases acquired on a fourth scanner from a manufacturer not represented in the training data. All data are provided in NIfTI format without preprocessing, and some cases contain randomly missing modalities. Our objective is to classify patients into four pathological stages (S1–S4), along with two binary subtasks:

1. Cirrhosis Detection (S1–S3 vs. S4)
2. Substantial Fibrosis Detection (S1 vs. S2–S4)

The training dataset comprising 360 patients was initially split into an internal training set and an internal validation set with a ratio of 0.75:0.25. To ensure robust evaluation, we further applied a 4-fold cross-validation strategy and stratified sampling was employed to guarantee that samples from each fibrosis stage were proportionally represented in each fold, preserving the stage distribution across all subsets [13].

### 2.2 Model Architecture

We adopt modality-specific and modality-correlated encoders from mmFormer, keeping their architectures largely consistent with the original design [7]. Our modifications center on a modality-compensation mechanism and a classification-oriented redesign.

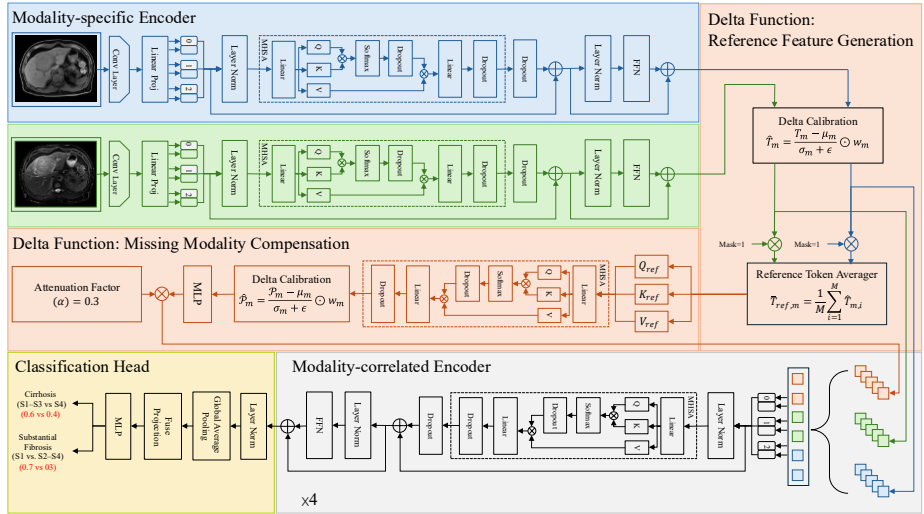

**Fig. 1.** The overall architecture of the model includes mmFormer's original modality-specific encoder, modality-correlated encoder, Delta Function for generating proxy features, and classification head [7].

### 2.3 Modality-specific Encoder

Modality-specific encoder integrates a convolutional encoder with an intra-modality Transformer: local patterns are captured by the convolutional encoder; the produced feature maps are then processed by an intra-modality Transformer to learn long-range dependencies. For each available modality $m \in \{T1WI, T2WI, DWI\}$, it is processed by an identical modality-specific encoder with independent parameters from input volume $X_m \in \mathbb{R}^{1 \times D \times H \times W}$.

**Convolutional Encoder.** The convolutional encoder $\mathcal{F}_m^{conv}$ stacks five residual 3D convolutional stages. Each stage begins with an entry 3D convolution. Stage 1 keeps unit stride, whereas stages 2 to 5 begin with an entry normalized convolution of stride 2 with reflect padding. At stage $s$, the cumulative down-sampling factor is $2^{s-1}$; thus, at the top stage $s = 5$, it is 16. Thus, the top-stage spatial dimension is $\frac{D}{16} \times \frac{H}{16} \times \frac{W}{16}$. Within each stage, we apply two normalized convolutional units. Each unit consists of instance normalization, LeakyReLU, and dropout. The unit output is added to the stage input via an identity skip connection. Denoting the base width by $C_{base} = 8$, the channel schedule across stages follows

$$C_s = 2^{s-1} C_{base}, \qquad s \in \{1,2,3,4,5\}. \tag{1}$$

The resulting local-context feature maps are given by

$$F_m^{local} = \mathcal{F}^{conv}(X_m; \theta_m^{conv}), \qquad F_m^{local} \in \mathbb{R}^{C_5 \times \frac{D}{16} \times \frac{H}{16} \times \frac{W}{16}}. \tag{2}$$

The top-stage feature is then projected to a fixed embedding dimension and flattened into tokens for the intra-modality Transformer [7].

**Intra-modality Transformer.** To capture long-range dependencies within each modality, we use an intra-modality Transformer to explicitly model global context. For modality $m$, the branch tokenizes features and applies pre-norm residual blocks of multi-head self-attention (MSA) and a position-wise feed-forward network (FFN). We apply a 1×1×1 convolution acting as a linear projection to obtain $C_t = 256$ channels and flatten voxels into a token sequence:

$$T_m^{(0)} = vec(Conv_{1 \times 1 \times 1}(F_m^{local})), \qquad T_m^{(0)} \in \mathbb{R}^{N \times C_t}, \tag{3}$$

where $N = D'H'W'$, and $C_t$ is embedding size. A fixed sinusoidal positional encoding $P_m \in \mathbb{R}^{N \times C_t}$ is used and added at every layer of the Transformer. At layer $l$, positions are added as

$$\tilde{T}_m^{(l)} = T_m^{(l-1)} + P_m. \tag{4}$$

With $Q_m^{(l)} = LN(\tilde{T}_m^{(l)})W_{Q,m}^{(l)}$, $K_m^{(l)} = LN(\tilde{T}_m^{(l)})W_{K,m}^{(l)}$, $V_m^{(l)} = LN(\tilde{T}_m^{(l)})W_{V,m}^{(l)}$, the output of MSA is

$$head_{m,i}^{(l)} = softmax\left(\frac{Q_m^{(l)}(K_m^{(l)})^\top}{\sqrt{d_k}}\right) V_m^{(l)}, \tag{5}$$

$$MSA_m^{(l)} = [head_{m,1}^1; \ldots; head_{m,h}^{(l)}] W_{O,m}^{(l)}, \tag{6}$$

where $LN(\cdot)$ represents layer normalization and $h = 4$ denotes the number of attention heads. Each layer stacks MSA and FFN with residual connections and the token sequence with long-range dependencies from intra-modality Transformer is shown as

$$Z_m^{(l)} = MSA_m^{(l)}(\tilde{T}_m^{(l)}) + \tilde{T}_m^{(l)}, \tag{7}$$

$$T_m^{(L)} = FFN_m^{(l)}\left(LN(Z_m^{(l)})\right) + Z_m^{(l)}, \tag{8}$$

where FFN is a two-layer position-wise perceptron with GELU activation and dropout. All parameters are modality-specific and not shared across modalities, enabling each Transformer to enhance modality-unique textures and lesion edges. It also produces stable, global-context features for the later cross-modal fusion and the classifier [7].

### 2.4 Delta Function

**Modality Mask.** We construct a binary modality-availability mask for each sample, where 1 indicates that the modality is present and participates in computation, and 0 indicates it is missing. When the modality branch is deemed "missing," the subsequent compensation module synthesizes proxy features to replace the missing modality [14].

**Delta Calibration.** To address cross-modality distribution mismatch in multimodal MRI, we are inspired by Feature-wise Linear Modulation (FiLM) and Adaptive Instance Normalization (AdaIN) and apply a learnable affine calibration to each modality's token sequence, which performs learnable mean-variance alignment and reliability-aware reweighting [15,16]. The formula can be shown as:

$$\hat{T}_m = \frac{T_m - \mu_m}{\sigma_m + \epsilon} \odot w_m, \tag{9}$$

where $\mu_m, \sigma_m, w_m$ are learnable per-channel mean, scale, and reliability weights, respectively. $\epsilon = 1 \times 10^{-8}$ ensures numerical stability. This approach maps different modality features to a shared space, improving cross-modal fusion and enabling better multimodal processing. After calibration, the modality-wise tokens have two purposes:

1. Calibrated tokens as valid modality are fed into the subsequent cross-modal fusion.
2. Calibrated tokens are aggregated as reference tokens to synthesize proxy features.

**Reference Token Averager.** To create a unified representation for missing modality compensation, we build a reference token by averaging features from all available modalities. The reference token is computed as:

$$\bar{T}_{ref,m} = \frac{1}{M} \sum_{i=1}^{M} \hat{T}_{m,i}, \tag{10}$$

where $M$ is the number of available modalities.

**Proxy Feature Compensation.** After generating the reference sequence, we apply another MSA to it to capture global context; therefore $\bar{T}_{ref,m} = Q_{ref} = K_{ref} = V_{ref}$ and preliminary proxy features $\mathcal{P}_m$ can be expressed as

$$\mathcal{P}_m = MSA(\bar{T}_{ref,m}, \bar{T}_{ref,m}, \bar{T}_{ref,m}). \tag{11}$$

Subsequently, preliminary proxy features are passed through affine calibration again to obtain

$$\hat{\mathcal{P}}_m = \frac{\mathcal{P}_m - \mu_m}{\sigma_m + \epsilon} \odot w_m, \tag{12}$$

ensuring the resulting target-modality proxy features are more robust [15]. Finally, the preliminary proxy features are refined by FFN, and a fixed attenuation factor $\alpha = 0.3$ is applied to attenuate their relative contribution and suppress noise from synthesized information during subsequent fusion [17]. The final proxy feature is obtained as

$$T_{proxy,m} = \alpha \times FFN(\hat{\mathcal{P}}_m). \tag{13}$$

### 2.5   Modality-correlated Encoder

Given the valid and proxy token sequences, we apply spatial alignment and standardize the token length by truncating longer sequences to the shortest length. All tokens are subsequently concatenated along token axis in a fixed order to yield the final sequence

$$T_c^{(0)} = [T_1, T_2, \cdots, T_M], \qquad T_c \in \mathbb{R}^{(M \times N_t) \times C_t}, \tag{14}$$

where $M$ is the number of available modalities, $T_c$ is concatenated token, and $N_t$ is the number of tokens per modality after alignment. This encoder stacks 4 Transformer blocks. At each layer, we add a fixed sinusoidal positional encoding

$$\tilde{T}_c^{(l)} = T_c^{(l-1)} + P_c, \tag{15}$$

and apply MSA and FFN with residual connections again [7]:

$$Z_c^{(l)} = MSA^{(l)}(\tilde{T}_c^{(l)}) + \tilde{T}_c^{(l)}, \tag{16}$$

$$T_c^{(L)} = FFN^{(l)}\left(LN(Z_c^{(l)})\right) + Z_c^{(l)}. \tag{17}$$

### 2.6   Classification Head

An external layer normalization is applied to the token sequence before pooling. Global average pooling (GAP) over the token yields a 256-dimensional representation, which is linearly projected with dropout to fused embedding. Also, a two-layer multi-layer perceptron (MLP) and the final linear projection layer are applied to output logits.

### 2.7   Model Ensemble

We employ four-fold cross-validation to train four homogeneous base models. At inference, the same sample is passed through all four models to obtain four logits, which are converted to class probabilities via SoftMax. We then perform soft voting by averaging the unweighted model probabilities for each class. Model ensemble improves robustness and generalization across splits [18].

## 3   Experiment and Results

Both subtasks follow the same experimental procedures and use the same equipment.

### 3.1 Experiment Setting

**Image Preprocessing.** For both training and inference, all images are reoriented to right–anterior–superior (RAS) and resampled to 1.5×1.5×3.0 mm voxel spacing, improving concordance and reducing sensitivity to pixel-spacing changes [19]. Then, intensities are normalized per volume by mapping the 1st–99th percentiles to [0,1] with clipping. Volumes are resized to 200×200×64 via symmetric padding or centered cropping, after which the first 20 axial slices are removed to improve target visibility, yielding a final input size of 200×200×44. Missing modalities are substituted with zero-filled volumes matching the target dimensions. To accelerate training, the preprocessed modality data and corresponding modality masks are packaged into NPZ files. Finally, stage-wise data augmentations are applied to the training set: its strength scales with sample size and combines geometric (flip/rotate/zoom) and intensity perturbations (scaling, noise, contrast, smoothing, shift).

**Training Hyperparameters.** The training script uses the AdamW optimizer with a learning rate of $1 \times 10^{-4}$ and weight decay of $1 \times 10^{-3}$, with cross-entropy loss. A cosine-annealing schedule is applied for up to 100 epochs with a minimum learning rate of $1 \times 10^{-6}$. Early stopping is triggered by validation loss with a patience of 30 epochs while continually saving the best checkpoint. Data are loaded with a batch size of 8. Training and inference are conducted on an NVIDIA A100 GPU with 40 GB VRAM.

### 3.2 Evaluation Metrics

We report results under the official metrics defined by the organizers, including accuracy (ACC) and AUC. Concretely, accuracy is the proportion of correct predictions among all evaluated samples [20]. As the decision threshold varies, the receiver operating characteristics (ROC) curve maps sensitivity against the false-positive rate; the area under receiver operating characteristics (AUROC) yields a threshold-independent discrimination score [21].

### 3.3 Model Evaluation

Table 1. Comparison between prior models and proposed model on in-distribution (ID) vendors.

| Method | Cirrhosis Detection (S1–S3 vs. S4) | | Substantial Fibrosis Detection (S1 vs. S2–S4) | |
| --- | --- | --- | --- | --- |
| | ACC (%) | AUC (%) | ACC (%) | AUC (%) |
| CE | 60.00 | 74.98 | 73.33 | 61.11 |
| BCE | **70.00** | 77.90 | 70.00 | 66.67 |
| BCE with CSL | 63.33 | **80.83** | 73.33 | 68.52 |
| Reg | **70.00** | 71.81 | 70.00 | **76.39** |
| CE with CSL | **70.00** | 76.83 | **80.00** | 75.23 |
| Proposed model (ID) | 66.67 | 71.73 | 74.17 | 68.48 |

As shown in Table 1, we compare our model against the as-reported external baselines from the LiFS task of CARE 2024, including CE, BCE, Reg, BCE with CSL, and CE with CSL[1] [22]. Without using any external data or relying on pre-segmentation or lesion localization, our approach maintains competitive performance overall. For Cirrhosis Detection, the model achieves accuracy of 66.67% and AUC of 71.73%; these are 3.33% and 9.10% lower than the best results, respectively, but improved 6.67% in accuracy over the CE baseline. For Substantial Fibrosis Detection, the model achieves accuracy of 74.17% and AUC of 68.48%, ranking second in accuracy and fourth in AUC, but only 0.04% behind the third-place method (68.52%). Compared to the CE baseline, this method improves accuracy by 0.84% and AUC by 7.37% on this task. Overall, without the need for additional segmentation or external data, this method demonstrates stable discriminative ability and good robustness on both tasks.

**Table 2.** Comparison of proposed model on ID vendors and out-of-distribution (OOD) vendors.

| Method | Cirrhosis Detection (S1–S3 vs. S4) | | Substantial Fibrosis Detection (S1 vs. S2–S4) | |
|---|---|---|---|---|
| | ACC (%) | AUC (%) | ACC (%) | AUC (%) |
| Proposed model (ID) | **66.67** | **71.73** | 74.17 | 68.48 |
| Proposed model (OOD) | 64.29 | 68.83 | **91.43** | 71.38 |

Under an OOD vendor, Table 2 shows that Cirrhosis Detection achieves accuracy of 64.29% and AUC of 68.83%, which are 2.38% and 2.90% lower than the in-distribution results, while Substantial Fibrosis Detection improves markedly to accuracy of 91.43% and AUC of 71.38%, gains of 17.26% and 2.90%, respectively, with 91.43% being the best among all evaluated methods.

## 4     Conclusion

Our study presents a multimodal Transformer based on mmFormer that integrates Delta-based missing-modality compensation, an improved classification head, and a four-fold ensembling strategy. The model shows strong adaptability to missing-modality scenarios in the LiFS task. On the in-distribution vendor split, it achieved accuracies of 66.67% and 74.17% for cirrhosis and substantial fibrosis detection, with corresponding AUCs of 71.73% and 68.48%, indicating competitive discrimination and robustness. Future work will focus on enhancing cross-center and cross-vendor generalization, with systematic external validation.

**Acknowledgments.** We thank the CARE 2025 organizers for providing access to the LiQA dataset used in this study [10-12]

**Disclosure of Interests.** The authors declare that they have no competing interests.

---

[1]     Abbreviations: CE for cross-entropy; BCE for binary cross-entropy; CSL for class activation map–segmentation map loss; Reg for regression [22].

10      Z. Zhang, J. Wang, and L. Zhang16. Huang, X., Belongie, S.: Arbitrary style transfer in real-time with adaptive instance normalization. In: Proceedings of the IEEE International Conference on Computer Vision (ICCV), pp. 1501–1510. IEEE, Venice (2017)
17. Oktay, O., Schlemper, J., Le Folgoc, L., Lee, M., Heinrich, M., Misawa, K., et al.: Attention U-Net: learning where to look for the pancreas. arXiv:1804.03999 (2018)
18. Ganaie, M.A., Hu, M., Malik, A.K., Tanveer, M., Suganthan, P.N.: Ensemble deep learning: a review. Engineering Applications of Artificial Intelligence 115, 105151 (2022)
19. Park, S.H., Lim, H., Bae, B.K., Hahm, M.H., Chong, G.O., Jeong, S.Y., Kim, J.C.: Robustness of magnetic resonance radiomic features to pixel size resampling and interpolation in patients with cervical cancer. Cancer Imaging 21(1), 19 (2021)
20. Grandini, M., Bagli, E., Visani, G.: Metrics for multi-class classification: an overview. arXiv:2008.05756 (2020)
21. Huang, J., Ling, C.X.: Using AUC and accuracy in evaluating learning algorithms. IEEE Transactions on Knowledge and Data Engineering 17(3), 299–310 (2005)
22. Zhang, H., Zhang, M., You, X., Gu, Y., Yang, G.-Z.: Computing assessment for liver fibrosis staging using real-world MR images. In: Zhuang, X., Ding, W., Wu, F., Gao, S., Li, L., Wang, S. (eds.) Comprehensive Analysis and Computing of Real-World Medical Images. CARE 2024, LNCS, vol. 15548, pp. 87–95. Springer, Cham (2025)